\documentclass[conference]{IEEEtran}
\usepackage{cite}
\usepackage{amsmath,amsfonts}
\usepackage{algorithmic}
\usepackage{graphicx}
\usepackage{subfig}
\usepackage{float}
\usepackage{multirow}
\usepackage{array}
\usepackage{textcomp}

\usepackage[hyperindex,breaklinks,hidelinks]{hyperref}
\hypersetup{%
   pdftitle={Improving Slow Transfer Predictions: Generative Methods Compared},%
   pdfauthor={Jacob Taegon Kim, Alex Sim, Kesheng Wu, Jinoh Kim},
   pdfkeywords={class imbalance, data augmentation, generative sampling, CTGAN, prediction, machine learning},
   colorlinks=true,     
   linkcolor=blue,      
   citecolor=blue,      
   filecolor=blue,      
   urlcolor=blue        
}

\graphicspath{{figs}}

\usepackage{xcolor}
\def\BibTeX{{\rm B\kern-.05em{\sc i\kern-.025em b}\kern-.08em
    T\kern-.1667em\lower.7ex\hbox{E}\kern-.125emX}}

\usepackage{booktabs}

\begin{document}
\title{Improving Slow Transfer Predictions: Generative Methods Compared}

\author{\IEEEauthorblockN{Jacob Taegon Kim}
\IEEEauthorblockA{
University of California, Berkeley\\
Berkeley, CA, USA\\
jacobkim@berkeley.edu}
\and
\IEEEauthorblockN{Alex Sim, Kesheng Wu}
\IEEEauthorblockA{
Lawrence Berkeley National Laboratory\\
Berkeley, CA, USA\\
\{asim, kwu\}@lbl.gov}
\and
\IEEEauthorblockN{Jinoh Kim}
\IEEEauthorblockA{
Texas A\&M University\\
Commerce, TX, USA\\
jinoh.kim@tamuc.edu}
}

\maketitle

\begin{abstract}
Monitoring data transfer performance is a crucial task in scientific computing networks. By predicting performance early in the communication phase, potentially sluggish transfers can be identified and selectively monitored, optimizing network usage and overall performance. A key bottleneck to improving the predictive power of machine learning (ML) models in this context is the issue of class imbalance. This project focuses on addressing the class imbalance problem to enhance the accuracy of performance predictions. In this study, we analyze and compare various augmentation strategies, including traditional oversampling methods and generative techniques. Additionally, we adjust the class imbalance ratios in training datasets to evaluate their impact on model performance. While augmentation may improve performance, as the imbalance ratio increases, the performance does not significantly improve. We conclude that even the most advanced technique, such as CTGAN, does not significantly improve over simple stratified sampling.
\end{abstract}


\section{Introduction}

Monitoring data transfer performance is one of the essential tasks in scientific computing networks~\cite{lu2005characterizing}.
While keeping track of large transfers (“elephant flows”) would be crucial considering their high bandwidth share and thus their impact on overall network performance, understanding the characteristics of “slow” transfers would have significance in regard to user experience and network operations.
For example, any possibility of (predicted) sluggish activities can be informed to the user in an early stage of the communication. This early notification would be beneficial since the user has a chance to consider an alternative way, and at the same time, the network can save the usage of its resources from being wasted by not-in-progress connections. In that regard, predicting data transfer performance before launching actual transfer would be essential for optimizing network usage and performance. With the predicted performance, a transfer that could be sluggish can be predicted in the initial communication phase and  can be monitored in a selective fashion rather than keeping track of the entire every single transfer. 

Unfortunately, however, it is highly challenging to perform the monitoring task in a per-transfer manner.
One of the challenges is the class imbalance concern that often adversely impact on prediction performance~\cite{4410370}.
In fact, class imbalance appears frequently in real-world scenarios, and is often the bottleneck to the performance of machine learning (ML) models.
For instance, the data transfer log this study refers to contain very small slow transfers in quantity (0.0002\% for slow vs. 0.9998\% for non-slow transfers).
This study focuses on a class imbalance issue to understand, characterize, and resolve the imbalance problem for enhancing prediction performance. 

When dealing with class imbalance, well-known approaches are undersampling and oversampling techniques. While undersampling addresses class imbalance issue, it will lead to insufficient datasets as well as creating imperfect representation of the datasets. On the other hand, oversampling may cause overfitting, and arguably worse, provide bunch of noise. 
On our previous work in~\cite{Shao:2022:slow}, we have studied different techniques for undersampling such as stratified sampling.
In this study, we take a data augmentation approach to see its feasibility on slow transfer prediction. 

Data augmentation is often considered for mitigating the class imbalance problem, which adds up minority class samples via, for example, statistical methods (e.g., Synthetic Minority Over-sampling Technique (SMOTE)~\cite{SMOTE} and Adaptive Synthetic Sampling (ADASYN)~\cite{ADAYSN}) or neural network-based generative tools (e.g., Generative Adversarial Network (GAN)~\cite{GAN} and its variants such as Conditional Tabular GAN (CTGAN)~\cite{CTGAN}).
In this study, we investigate how generating synthetic minor class samples could help improve the overall predicting accuracy in the context of slow-transfer prediction.
We experiment different data augmentation techniques to evaluate their impact on performance, and present our augmentation method based on CTGAN, which outperforms the existing augmentation tools, but not significantly better than simple stratified sampling.
This experimental results can be used for predicting transfer performance and classifying data transfers based on the prediction (e.g., slow or elephant flows) in scientific facilities.

\section{Related Work}
    \label{sec:related}

Monitoring network traffic is essential in network operations and management for detecting anomalous events and estimating network performance. In the realm of scientific computing, this task becomes even more critical due to the increasing demands of data-intensive exploration and computational activities. Identifying "elephant flows"—large data transfers that consume significant network capacity—has been a focal point in prior research. For instance, Basat et al.\cite{basat2017optimal} introduced an algorithm that estimates the traffic volume of individual flows to detect elephant flows based on total byte counts. They utilized two hash tables to record counters associated with flow IDs from packet traces, enabling the detection of flows exceeding a predefined threshold. Similarly, Chhabra et al.\cite{chhabra2017classifying} approached the classification of elephant (large transfer) and mice (small transfer) flows using an unsupervised Gaussian Mixture Model clustering based on NetFlow data. While these studies emphasize the importance of identifying large flows, our research focuses on predicting slow connections, which can significantly impact the performance of data-intensive scientific applications.

Several studies have analyzed {\tt tstat} data to improve network performance monitoring and anomaly detection. Syal et al.\cite{syal2019automatic} presented a classification mechanism to detect low-throughput time intervals. Their approach involved a two-phase process: first, assigning binary labels (anomalous or not) to each time window, and second, constructing a supervised learning model using these labels. Nakashima et al.\cite{nakashima2020evaluation} evaluated various deep learning models—including Multilayer Perceptrons, Convolutional Neural Networks, Gated Recurrent Units, and Long Short-Term Memory networks—to predict network performance in terms of aggregated throughput per time interval. While these studies leveraged {\tt tstat} data with a focus on time-windowed analysis, our study distinguishes itself by aiming for connection-level prediction, providing a more granular understanding of network performance issues.

Addressing class imbalance is a critical challenge in machine learning, especially in network anomaly detection where anomalous events are rare compared to normal traffic. Data augmentation techniques have been widely employed to mitigate this issue by artificially increasing the representation of minority classes. Methods such as SMOTE and its variants have been effectively used in various domains to balance datasets~\cite{ELREEDY201932, fernandez2018smote}. However, these methods may not fully capture the complex distributions inherent in structured network data. On the other hand, generative models, especially GAN, have shown significant potential in generating realistic data and have been adapted to various fields beyond image synthesis. Despite originating from image processing, GANs have been increasingly applied to structured data like tabular datasets, though challenges remain due to the inherent differences in data types~\cite{zhao21}. Recent studies have begun to explore the use of GANs for oversampling in structured data domains. For example, Loey et al.\cite{loey} applied CGAN to COVID-19 detection in chest CT scan image. Bourou et al. \cite{info12090375} utilized TGAN and CTGAN to create synthetic Intrusion Detection Systems (IDS) datasets to enhance network security tools. Wang et al.~\cite{s22145243} applied CTGAN to traffic data, creating a new model CTTGAN for detecting malicious network traffic.

\section{Data Augmentation for Data Transfer Throughput Prediction}

Standard oversampling methods like SMOTE and their variations generate synthetic samples along the line segment that joins minority class samples. Therefore, these approaches are based on local information, rather than on the overall minority class distribution. Since the separation between majority and minority class clusters is not often clear, noisy samples may be generated.
To avoid this, various undersampling (filtering) variations have been tried such as SMOTE-ENN~\cite{ENN}, and Borderline-SMOTE~\cite{Borderline-SMOTE}. A direct approach to the data augmentation process would be the use of a generative model that captures the actual data distribution.
In this section, we describe oversampling and generative methods for augmenting slow data transfer records observed much less in real scientific computing environments.

\subsection{Oversampling Methods}

\subsubsection{SMOTE}
SMOTE generates samples by interpolating between randomly selected minority samples and $k$ nearest neighbors. This technique improves on the issue of overfitting as it generates a random sample that is not a duplicate of an existing minority sample unlike random oversampling. The algorithm is as follows: 1) randomly select a sample from minority samples, 2) use KNN to select $k$ nearest neighbors around minority sample, 3) interpolate between these points to create a random sample. Then, the algorithm repeats this process until the desired number of synthetic samples.
\vspace{-1mm} 
\begin{align}
\begin{split}
    x_{\text{smote}} = x_i + (\hat{x_i} - x_i) \times \delta, \\
    \delta \in [0,1], \\
    i = 1,2, ... , k
\end{split}
\end{align}

where $x_i$ is random sample in minority class, and $\hat{x_i}$ is random KNN

\subsubsection{Borderline SMOTE}
Naive SMOTE might generate too much noise. Hence, a Borderline SMOTE is designed to be an improvement to SMOTE by avoiding excessive noise generation. It generates synthetic samples near the decision boundary (borderline) between classes rather than randomly in the minority class space. The algorithms is as follows: 1) Find instances of the minority class that have a mix of majority class neighbors (boundary between majority and minority space), 2) among the borderline instances, identify those that are misclassified using a $k$-nearest neighbor classifier, then 3) generate synthetic samples by picking a random minority class neighbor, and interpolating between the instance and its neighbor.

\subsubsection{SMOTE ENN}
Start SMOTE process. After oversampling is finished, start Edited Nearest Neighbor undersampling. Find the KNN of the all samples. If the majority class of the sample’s KNN and the sample’s class is different, then the sample and its KNN are deleted from the dataset.

\subsubsection{SMOTE Tomek-Link}
Start SMOTE process. After oversampling is finished, start Tomek-Link undersampling. Tomek-Link is one of a modification from Condensed Nearest Neighbors, which is defined as selecting the pair of observation $a$ and $b$ such that they are each other's nearest neighbor and they do not belong to the same class. After choosing a random data from the majority class, if the random data’s nearest neighbor is the data from the minority class (i.e. Tomek Link), then remove the Tomek Link.

\subsubsection{ADASYN}
ADASYN is similar to SMOTE but it generates different number of samples depending on an estimate of the local distribution of the class to be oversampled i.e. ADASYN adapts to the underlying data distribution by generating synthetic samples in regions where the minority class is underrepresented. The algorithm is as follows: 1) For each sample in the minority class, calculate its $k$ nearest neighbors, 2) for each minority class sample, compute a ratio (density distribution) that dictates how many synthetic samples will be generated for that sample, depending on how many majority samples are within the KNN, then 3) generate synthetic samples for each minority class sample based on the computed ratio:
\vspace{-5mm} 
\begin{align}
\begin{split}
    r_i = \frac{\Delta_i}{K}, \\
    i = 1,2, ..., C_{\text{min}}
\end{split}
\end{align}
after normalizing it, where $\Delta_i$ is the number of samples in the $K$ nearest neighbors of $x_i \in C_{\text{min}}$ that belong to the majority class.

\subsection{Generative Methods}
\subsubsection{GAN}
GAN is a deep-learning unsupervised model that is based on the idea of game theory, in which a generator $G$ and a discriminator $D$ are trying to outsmart each other. The objective of the generator is to confuse the discriminator. The objective of the discriminator is to distinguish the instances coming from the generator and the instances coming from the original dataset. The generative model $G$, defined as $G: Z \rightarrow X$ where $Z$ is the noise space of arbitrary dimension that corresponds to a hyperparameter and $X$ is the data space, aims to capture the data distribution. The discriminative model, defined as $D: X \rightarrow [0, 1]$, estimates the probability that a sample came from the data distribution rather than $G$. Then the objective of the game is given as:
\vspace{-1mm} 
\begin{align}
\begin{split}
    \min_G \max_D V(D,G) = \\
    E_{x \sim p_{\text{data}}(x)} [\log D(x)] + E_{z \sim p_z(z)} [\log (1-D(G(z))]
\end{split}
\end{align}

In the equation, $p_{\text{data}}(x)$ and $p_z(z)$ refer to the distribution of real data and distribution of input noise variable, respectively. The generator $G$ aims to produce data $G(z)$ that the discriminator $D$ cannot distinguish from real data $x$, while $D$ strives to correctly differentiate between real and generated data. Through iterative training, $G$ and $D$ optimize this value function, leading $G$ to generate increasingly realistic data over time.

\subsubsection{CTGAN}
Traditional GAN algorithms have shown impressive capabilities in learning original images and generating synthetic ones under various conditions. However, their application to structured data is limited due to challenges such as diverse tabular data types, non-Gaussian distributions, multimodal data, sparse matrices from one-hot encoding, and highly imbalanced categorical variables. To tackle these limitations, CTGAN was developed. This model integrates conditional-GAN and tabular-GAN algorithms, addressing the common GAN issue of not effectively learning sparse categories by including conditions in the learning process. CTGAN employs mode-specific normalization and training-by-sampling to manage the issues of multimodal and non-Gaussian data distributions. During training, it normalizes numerical data using the variational Gaussian mixture, and post-training, it converts generated data back to the original scale.


\begin{table*}[!tb]
\vspace{-5mm} 
\centering
\footnotesize
\caption{Datasets Information with Imbalance Ratio (IR)}%
\begin{tabular}{llll}
\toprule
\textbf{Data} & \textbf{Year} & \textbf{Description} & \textbf{IR (\# Samples)}\\
\midrule
\texttt{train1} & \texttt{2021} & Random sampling & \texttt{1:1000}\\
\texttt{train2} & \texttt{2021} & Keep the entire slow transfers and randomly sample the same number of normal transfers & \texttt{3000:3000} \\
\texttt{train3} & \texttt{2021} & Keep half of slow transfers and randomly sample twice the size of the slow transfers for normal transfers & \texttt{1500:3000} \\
\texttt{train4} & \texttt{2021} & Keep the entire slow transfers and sample normal transfers twice the size of the slow transfers & \texttt{3000:6000} \\
\texttt{train5} & \texttt{2021} & Sample 1,000 slow transfers and sample 10,000 normal transfers & \texttt{1000:10000} \\
\texttt{test2} & \texttt{2022} & Keep the entire slow transfers and randomly sample the same number of normal transfers & \texttt{1:1}\\
\bottomrule
\end{tabular}
\label{tab:datasets}
\vspace{-5mm} 
\end{table*}

\section{Experimental Setting}
The datasets are provided from \texttt{tstat}, a logging tool for network traffic flows which is deployed on data transfer nodes (DTN) in National Energy Research Scientific Computing Center (NERSC\footnote{NERSC, https://www.nersc.gov/}).
For this study, we use all the data collected in 2021 as the training set and all the data from 2022 as the testing set.

For all training datasets, we divided into two sub-classes: minority $C_{\text{min}}$ and majority $C_{\text{maj}}$. For minority class $C_{\text{min}}$, we applied augmentation methods to balance the class imbalance ratio. Class imbalance ratio, as we define is simply the proportion of minority class to majority class, $IR = |C_{\text{min}}| / |C_{\text{maj}}|$. In Section~\ref{sec:results}, we use this imbalance ratio to assess the predictive power augmented in the model.

In this study, our main augmentation methods are 1) oversampling methods and 2) generative methods. In oversampling, the idea is to interpolate between a sample $x_i \in C_{\text{min}}$ and its neighbors, which then may be followed by filtering and editing. SMOTE-ENN and SMOTE-Tomek Links are latter examples. First, we applied SMOTE to minority class of training datasets, and trained models on this new augmented datasets. Similarly, we applied Borderline-SMOTE, SMOTE-ENN, SMOTE-Tomek Links, and ADASYN to the minority samples and compared the results. SMOTE-Tomek Links and SMOTE-ENN are hybrid of oversampling and undersampling methods, where after augmentation is done, it needs to satisfy criterion to be validated into resulting training sets. Hence, the resulting dataset is smaller.

We chose \texttt{train2-test2} as a Baseline, then compared the results of different data augmentation methods. Hyperparameters for Random Forest and XGBoost was fine-tuned using GridSearchCV and \texttt{train2-test2} data sets as a baseline model.

\subsection{Evaluation Metrics}

To thoroughly evaluate our models, we utilize a combination of quantitative and qualitative metrics that not only measure performance but also provide insights into the data's structure and distribution. This approach ensures a comprehensive understanding of how well our models perform and how closely synthetic data resembles real data.

\subsubsection{Quantitative Metrics}
    {\bf F1-Score:} Given the imbalance in our dataset—with slow connections being a minority—we avoid relying on simple accuracy, which could be misleading. Instead, we use the F1-Score, the harmonic mean of Precision and Recall, to better assess the model's ability to correctly identify slow connections. This metric provides a balanced evaluation of performance, particularly in cases of class imbalance. We also employ 10-fold cross-validation to ensure that our results are robust and not due to random chance.
    {\bf Kolmogorov–Smirnov (KS) Test:} To determine how closely our synthetic data matches the real data, we use the KS Test. This statistical test measures the maximum difference between the cumulative distributions of two datasets. By applying the KS Test, we can quantitatively assess the similarity in distributions, which is crucial for validating the effectiveness of our data generation methods.

\subsubsection{Qualitative Metrics}

For a deeper understanding of the data and to interpret the results more intuitively, we incorporate visualization techniques that allow us to observe the structural relationships within the data. We utilize t-Distributed Stochastic Neighbor Embedding ({\bf t-SNE}) and Uniform Manifold Approximation and Projection ({\bf UMAP}) for dimensionality reduction and visualization. t-SNE is effective at preserving local structures in high-dimensional data, while UMAP aims to maintain both local and global structures. By comparing visualizations from both methods, we can qualitatively assess how well the synthetic data captures the underlying patterns of the real data.

\section{Results}
    \label{sec:results}
    
\begin{table}[tb!]
\vspace{1mm} 
\centering
\footnotesize
\caption{F1-Scores of various data augmentation methods. \texttt{Train3} was used as the baseline dataset for augmentation to balance the class ratio. Then, the performance was tested using \texttt{Test2} dataset.}%
\begin{tabular}{llll}
\toprule
\textbf{Dataset} & \textbf{Decision Tree} & \textbf{Random Forests} & \textbf{XGBoost} \\
\midrule
Train1 & 0.0 & 0.0 & 0.0 \\
Train2 & 0.882 & 0.895 & 0.896 \\
SMOTE & 0.884 & 0.9 & 0.895 \\
SMOTE Borderline & 0.903 & 0.897 & 0.894 \\
SMOTE Tomek & 0.884 & 0.9 & 0.896 \\
SMOTE ENN & 0.892 & 0.9 & 0.893 \\
ADASYN & 0.892 & 0.895 & 0.896 \\
GAN & 0.902 & 0.871 & 0.869 \\
CTGAN & 0.878 & 0.896 & 0.898 \\
\bottomrule
\end{tabular}
\label{tab:f1}
\vspace{-3mm}
\end{table}

\begin{figure}[tb!]
\vspace{-2.75mm}
\centering
\includegraphics[width=9cm]{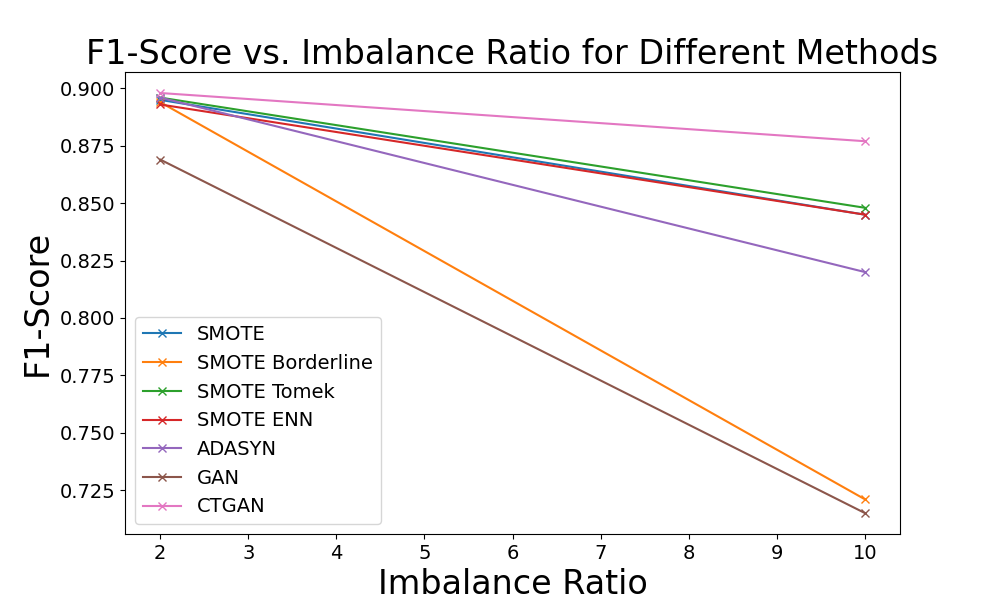}
\caption{Displaying the performance of data augmentation techniques across two different imbalance ratios 1:2 and 1:10. Each method is represented by a distinct line, with points marked at the respective imbalance ratios.}%
\label{fig:perf}
\end{figure}

For traditional data augmentation methods, we selected \texttt{train2-test2} as a baseline model and compared the F1-Scores of each methods. The full datasets used for training are listed in Table~\ref{tab:datasets}. Figure~\ref{fig:perf} displays the F1-scores of various augmentation methods as the imbalance ratio varies on the training set. As the imbalance ratio increases, the performance of the model decreases. Some models like GAN and SMOTE-Borderline show a sharp decrease in performance compared to CTGAN. We observed that CTGAN is robust in the sense that it continued to perform reasonably well on datasets with larger imbalance ratios (\texttt{1:10}). Hence, we chose CTGAN to assess the quality of generated samples.

\begin{table}[tb!]
\centering
\footnotesize
\caption{Top 6 features based KS Score. KS Score uses the inverse of Kolmogorov-Smirnov statistic to determine the distance of two distribution for original minority and synthetic minority samples. Higher score indicates higher resemblance.}%
\begin{tabular}{p{4cm} p{2cm}}
\toprule
\textbf{Feature} & \textbf{KS Score} \\
\midrule
\texttt{size} & 0.917 \\
\texttt{prev\_tput} & 0.906 \\
\texttt{prev\_size} & 0.892 \\
\texttt{prev\_durat} & 0.869 \\
\texttt{prev\_rtt\_max} & 0.835 \\
\texttt{size\_ratio} & 0.819 \\
\bottomrule
\end{tabular}
\label{tab:top6}
\vspace{-5mm}
\end{table}

\begin{figure}[tb!]
\vspace{-2mm}
\includegraphics[width=9cm]{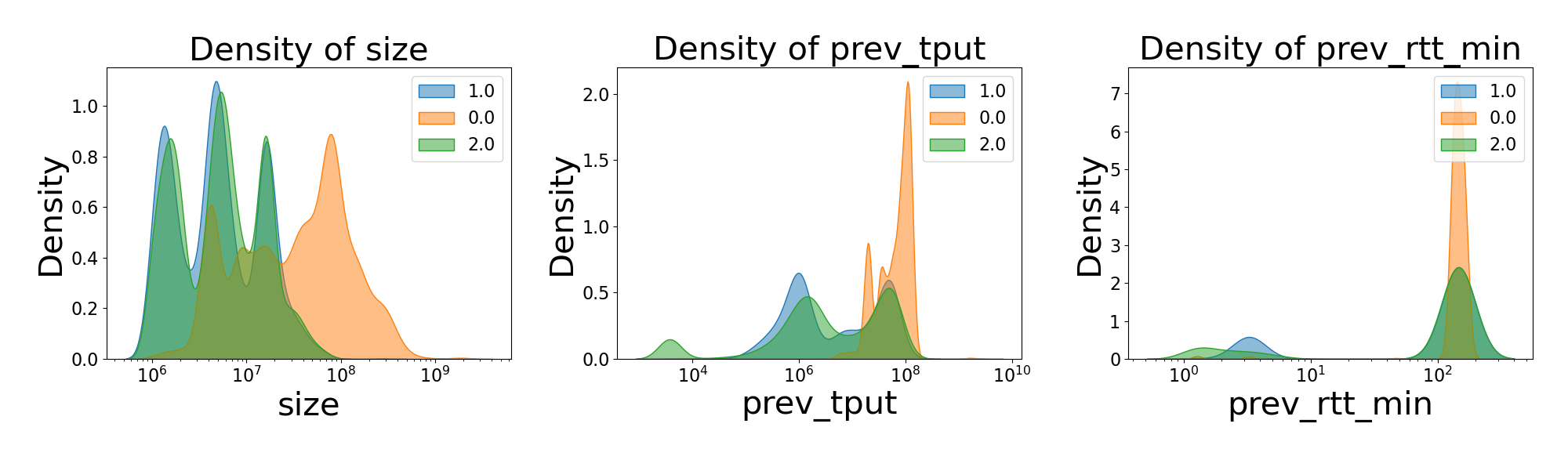}
\centering
\caption{Log histogram comparison of feature distributions for real and synthetic minority samples generated using CTGAN. The close overlap between the real and synthetic minority class distributions indicates a similarity in feature-level distribution.}%
\label{fig:histo}
\vspace{-3mm}
\end{figure}

To analyze the quality of synthetic samples, we compared the distribution of different classes (\texttt{0 = majority, 1 = minority, and 2 = synthetic minority}). For CTGAN, we only input minority samples from the training dataset and generated synthetic minority samples to balance the class ratio. We assessed the quality of the synthetically generated samples via two methods: 1) visualization and 2) KS-Test. After running 2000 epochs, we saw the convergence of the loss close to 0. The KS-Test scores result is shown in Table~\ref{tab:top6}. KS-Test compares two distributions and output similarity scores. For each features, the generated synthetic samples resemble true distribution of the feature if the score is high. We also compared the log distributions of the synthetic samples in Figure~\ref{fig:histo}. We first grouped by class labels and overlaid it in one plot to show the contrasts of the distributions. As you can see, the density of the synthetic minority samples closely align with that of original minority samples, validating the data augmentation process.

\begin{figure}[t!]
\includegraphics[width=8cm]{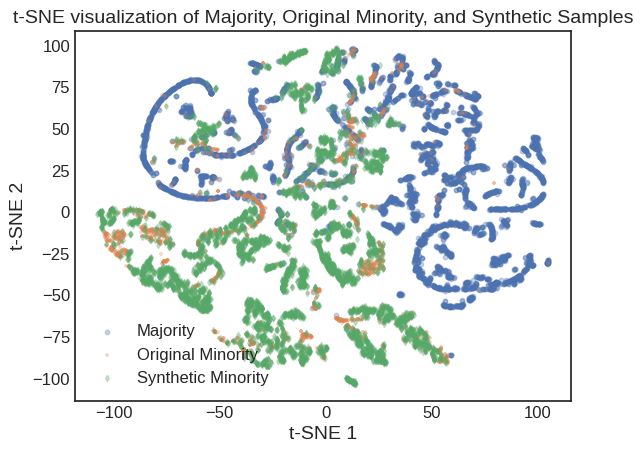}
\vspace{-2mm}
\centering
\caption{t-SNE showing non-overlapping clusters of original minority and synthetic minority samples.}%
\label{fig:tsne}
\vspace{-5mm}
\end{figure}

Though the log histogram show close overlaps between real and synthetic minority samples, it does not fully capture joint behavior of the features. Looking at Figure~\ref{fig:tsne}, the synthetic samples may replicate individual feature distributions but fail to capture the complex relationships between features present in the original data. Using t-SNE, which considers the high-dimensional relationships between data points, effectively capturing the joint distribution of features, non-overlapping clusters may indicate that the synthetic data differs from the original data when considering the combination of features. Hence, this may explain the performance of the model not significantly being improved. Table~\ref{tab:f1} shows the performance of different data augmentation methods used.

\section{Conclusions}

After comparing the results across different data augmentation techniques, we found that these methods do not consistently outperform stratified sampling. While the log histogram comparison of minority class distributions
showed a close overlap between real and synthetic samples, it does not take into account of joint distributions. A more sophisticated technique (t-SNE) 
revealed that synthetic minority samples behave differently from real minority samples. This suggests that even advanced techniques, like CTGAN, which performed well in generating visually similar distributions, are still unable to fully replicate the true characteristics of the minority class.
If the synthetic data does not accurately represent the joint feature relationships, the model may not benefit from the additional data, as it does not contribute to enhancing the model's ability to generalize to the minority class.
This discrepancy between t-SNE and log histogram is likely the reason for the lack of significant performance improvement over simpler methods like stratified sampling. Even with the best-performing augmentation techniques, such as CTGAN, the synthetic data does not fully capture the behavior of real minority data, which is reflected in the comparable performance between data augmentation techniques and stratified sampling.

\vspace{-4mm} 
\section*{Acknowledgements}
This work was supported by the Office of Advanced Scientific Computing Research, Office of Science, of the U.S. Department of Energy, under Contract No.~DE-AC02-05CH11231.  This work also used resources of the National Energy Research Scientific Computing Center (NERSC).

\bibliographystyle{unsrt}
\bibliography{main}

\end{document}